\documentclass[
twocolumn,
]{ceurart}

\usepackage{todonotes}
\usepackage[inline,shortlabels]{enumitem}
\usepackage{graphicx}
\usepackage{subfig}
\usepackage{hyperref}
\usepackage{url}
\usepackage[most]{tcolorbox}
\usepackage{fontawesome5}

\usepackage[nameinlink]{cleveref}
\crefname{figure}{Fig.}{Figs.}
\Crefname{figure}{Figure}{Figures}
\crefname{equation}{Eq.}{Eqs.}
\Crefname{equation}{Equation}{Equations}
\crefname{section}{§}{§§}
\Crefname{section}{Section}{Sections}
\crefname{table}{Table}{Tables}
\crefname{appendix}{Appendix}{Appendices}

\definecolor{skillcolor}{rgb}{1.0, 0.9, 0.8}
\definecolor{descriptioncolor}{rgb}{0.85, 0.91, 0.99}
\definecolor{sentencecolor}{rgb}{0.84, 0.91, 0.83}
\definecolor{othersentencecolor}{gray}{.9}

\newcommand{\experience}[1]{\textit{ex}_{#1}}
\newcommand{\esco}[1]{\textit{occ}_{#1}}

\usepackage{color}

\usepackage{color}

\usepackage{listings}
\newcommand{\eg}{e.g., }

\newcommand{\ie}{i.e., }

\begin{document}

\copyrightyear{2023}
\copyrightclause{Copyright for this paper by its authors.
  Use permitted under Creative Commons License Attribution 4.0
  International (CC BY 4.0).}

\conference{RecSys in HR'23: The 3rd Workshop on Recommender Systems for Human Resources, in conjunction with the 17th ACM Conference on Recommender Systems, September 18--22, 2023, Singapore, Singapore.}

\title{Career Path Prediction using Resume Representation Learning and Skill-based Matching}

 \author[1,2]{Jens-Joris~Decorte}[%
 email=jensjoris@techwolf.ai,
 url=https://www.techwolf.ai,
 ]
 \cormark[1]
 \address[1]{Ghent University -- imec,
   9052 Gent, Belgium}
 \address[2]{TechWolf,
   9000 Gent, Belgium}

 \author[2]{Jeroen~Van~Hautte}[%
 email=jeroen@techwolf.ai,
 url=https://www.techwolf.ai,
 ]

 \author[1]{Johannes~Deleu}[%
 email=johannes.deleu@ugent.be,
 ]

 \author[1]{Chris~Develder}[%
 email=chris.develder@ugent.be,
 ]

 \author[1]{Thomas~Demeester}[%
 email=thomas.demeester@ugent.be,
 ]

\cortext[1]{Corresponding author.}

\begin{abstract}
The impact of person-job fit on job satisfaction and performance is widely acknowledged, which highlights the importance of providing workers with next steps at the right time in their career.
This task of predicting the next step in a career is known as career path prediction, and has diverse applications such as turnover prevention and internal job mobility. 
Existing methods to career path prediction rely on large amounts of private career history data to model the interactions between job titles and companies.
We propose leveraging the unexplored textual descriptions that are part of work experience sections in resumes. 
We introduce a structured dataset of 2,164 anonymized career histories, annotated with ESCO occupation labels. 
Based on this dataset, we present a novel representation learning approach, CareerBERT, specifically designed for work history data.
We develop a skill-based model and a text-based model for career path prediction, which achieve 35.24\% and 39.61\% recall@10 respectively on our dataset.
Finally, we show that both approaches are complementary as a hybrid approach achieves the strongest result with 43.01\% recall@10.
\end{abstract}


\maketitle

\section{Introduction}

It is well-known that person-job fit has a positive impact on both job satisfaction and job performance~\cite{iqbal2012impact,chhabra2015person}.
Also, employment plays a large role in most people's lives and has an important impact on their well-being~\cite{de2018work}.
Thus, providing people with next steps at the right time in their career that are both inspiring and suited to their experience is important for job satisfaction, productivity and well-being of workers.
The task of predicting the next step in a career is known as career path prediction.
While it is closely related to job recommendation, career path prediction does not recommend specific job ads to candidates, but rather aims to predict the next role in an individual's career.
Such a role is typically characterized by a company name, job title and optional attributes such as salary or location.
Being able to predict next steps in individual's careers has many applications, ranging from turnover prevention to internal job mobility.

Common approaches to career path prediction rely on large amounts of career history data, and structure all career transitions into a large graph that contains both employers and job titles~\cite{zhang2021attentive,yamashita2022looking}.
Relying on only sparse features, such as job title and company names, necessitates large amounts of career trajectories in order to learn meaningful (graph) representations from them.
However, as such career data constitutes personal information, most research relies on closed datasets, often proprietary to a company. 
Hence, there is a lack of open datasets for the development and evaluation of career path prediction algorithms.

We believe that the career path prediction task can benefit from as of yet untapped unstructured data sources, \ie the free-form textual descriptions of past work experience in resumes.
Concretely, we propose a relatively small, anonymous dataset of textual career histories from resumes, enriched with structured occupation labels from a predefined ontology.
For the latter we adopt the European Skills, Competences, Qualifications and Occupations (ESCO)~\cite{ESCO}.
In this paper, we define the career path prediction task as follows: \textbf{given} a career history, \ie a sequence of experiences $(ex_1, ex_2, \dots, ex_{N-1})$ each having a title, description and their ESCO occupation labels $(occ_1, occ_2, \dots, occ_{N-1})$, \textbf{predict} the ESCO occupation label $occ_N$ of the held-out next experience.
We believe that by focusing on the prediction of the next occupation, such a system can help in recommending relevant next jobs or providing clarity on internal mobility at employers in the future.
Our main contributions are:
\begin{itemize}[leftmargin=*]
    \item We create, annotate and publish\footnote{\url{https://huggingface.co/datasets/jensjorisdecorte/anonymous-working-histories}} a dataset of 2,164 anonymous career histories across 24 different industries (\cref{sec:dataset}). The career histories are structured as a list of work experiences described in free-form text. Each experience is annotated with corresponding ESCO occupation.
    \item We show how the parallel information present in the textual career histories and in the occupation ontology provides opportunities to train a domain-specific text representation model (\cref{sec:models}) that can be used downstream for the career path prediction task, under a constrained dataset size.
    \item We show how the hybrid approach of combining text-based and skill-based prediction achieves the strongest results (\cref{sec:results}) for our task, thus demonstrating the value of injecting skill ontology information into the model (as opposed to using purely text-based models).
\end{itemize}

\section{Related Work}

\subsection{Resume Representation Learning}

We believe that expressive representations of resumes can benefit many HR-related tasks such as job recommendation and career path prediction.
Building qualitative resumes representations is challenging due to the semi-structured nature of resumes.
Resumes tend to contain similar sections, but within each section, the text is typically unstructured.
Current works on capturing resumes into more structured representations mostly focus on extracting only a subset of information present in resumes.
As a result, many approaches focus on just a subset of information present in resumes.
The Job2Vec model learns job title representations based on a graph of thousands of career paths in the IT and Finance~\cite{zhang2019job2vec}, but completely ignores the unstructured description linked to the experiences.
Another interesting work develops a similarity measure between careers (\textsc{SimCareers}) as a sequence alignment metric between sequences of positions~\cite{xu2014modeling}.
This work does use the unstructured summaries, but only after applying keyword extraction on them.\\
Only a minority of works aims to capture the full job position information and typically relies on matched pairs of resume text and job ads.
Examples of this are \cite{maheshwary2018matching} that train a siamese adaptation of convolutional neural network.
A more recent work uses contrastive learning of a sentence-transformer model between corresponding resume, job ad pairs~\cite{lavi2021consultantbert}.
The downside of these methods is effectively the need for a job recommendation dataset, which is hard to get access to, and may contain unexpected biases depending on how the data was gathered.\\
We propose a new way of learning expressive representations of textual career histories called \textsc{CareerBERT} without the need for resume, job pairs.
Instead, \textsc{CareerBERT} relies on textual career histories and their corresponding ESCO occupations labels only.

\subsection{Career Path Prediction}

In the field of career path prediction, large scale data from social networks (LinkedIn) has been an important source of information~\cite{liu2016fortune,yamashita2022looking,zhang2021attentive}.
An early work on career path prediction focused on four distinct career paths - software engineering, sales, consulting, and marketing~\cite{liu2016fortune}. They simplified these paths into four stages of seniority and normalized LinkedIn job titles accordingly for the prediction task. While the specific dataset is not publicly available, they extracted demographic, psycholinguistic, and topic-related features from social media content to enhance their predictions.
An extended approach that predicts multiple future job titles and company changes ahead, rather than just the next step was proposed by~\cite{yamashita2022looking}.
They utilized a proprietary dataset of 300,000 resumes, allowing them to delve deeper into career trajectory analysis, but only used job titles and companies as features for the task at hand.
Another approach to career path prediction uses an LSTM to represent both profile context and career path dynamics, leveraging a LinkedIn dataset to predict both the next company and job title~\cite{li2017nemo}.
Massive amounts of resumes (+459k) have been used to predict job mobility patterns using a heterogeneous company-position network constructed from the resumes' career trajectory data, providing insights into career transitions and progression~\cite{zhang2021attentive}.
All aforementioned methods rely on extensive collections of resumes and overlook the information embedded within the free-form text that is part of work experience sections.
In contrast, our work leverages this text to enable new methods, that do not require massive-scale datasets and interaction graphs, as the textual content could offer a richer context for understanding career progression. 

\section{Anonymous Career Path Dataset}
\label{sec:dataset}

We reuse the set of anonymous resumes~\cite{inoubli2022dgl4c, 10169980}, gathered from Kaggle,\footnote{\url{https://www.kaggle.com/datasets/snehaanbhawal/resume-dataset}} which contains 2,482 anonymous resumes, both in textual form and as pdf files.
These anonymized resumes were originally collected from an online portal, and are based on different profiles that applied on the platform to jobs from 24 different industries. 
In \cref{sec:creation}, we detail how we transformed these resumes into structured lists of experiences, each with their respective job title, experience summary, time period and ESCO occupation counterpart.
Then, \cref{sec:analytics} summarizes the main characteristics of the obtained dataset.

\subsection{Dataset Construction}
\label{sec:creation}

We parse structured career histories from the resumes in free-text form, as written by their authors.
Such career history is composed of a sequence of \emph{experiences} $\experience{1} \dots \experience{N}$, each defined as a \emph{title} and \emph{description} and the time period it covered.
The length $N$ of a career history may obviously differ across resumes.
We supplement each individual experience $\experience{i}$ with a corresponding ESCO occupation label $\esco{i}$.
Next, we detail how we extract the title and descriptions from the full-text resumes, as well as the process to obtain ESCO labels.

\paragraph{Extract experience section:}
Since we observed that the original dataset's text format lacks structure, presumably due to PDF or HTML parsing artefacts, we preprocess the data to restore paragraph segmentation.
Consecutive whitespaces were identified as suitable places to insert newlines, which reconstructs a readable format.
Since we are only interested in the professional experience listed in the resume, we want to skip all of the sections on ``education'', ``certifications'', ``projects'', ``skills'', ``publications'', ``awards'', ``personal information'', ``presentations'', etc.
We thus manually inspected the resumes in the dataset to identify the section titles used, and extract the \emph{experiences} of interest as the region in between one of the related experience headings\footnote{We found the following headings preceding the experiences of interest:
``experience'', ``professional experience'', ``work history'', ``work experience'', ``relevant experience'', ``relevant professional experience'', ``employment history'', ``employment \& experience''.} 
and the earliest subsequent section header.
The length of the thus selected sections on average amounts to 59\% of the original resume length.
We successfully processed 2,473 out of all 2,484 resumes, discarding the remaining 11 low quality resumes.

\paragraph{Structure working experiences:}
The obtained work experience sections list the different roles, often in chronological order.
Because the resumes are anonymized, experiences are annotated with general ``Company Name'' and ``City, State'' placeholders, which we thus neglect.
Each experience contains a job title (typically on a separate line) and a paragraph describing the 
respective responsibilities.
Finally, each experience contains the period in which it was performed, with start and end date (or ``current'').
The order in which title, period and description are mentioned varies across resumes, which makes it hard to uniformly define the separation (\eg as a regular expression) between each experience in the text.
Therefore, we rewrite the experience section in a uniform format using the \textsc{GPT-3.5} API (see \cref{app:gpt-prompt-for-reformatting}).
From that uniform text format, we then easily parse the text into a JSON structure combining the title, description, start and end date.
Finally only profiles with 2+ experiences are retrained, after which 2,164 career histories remain.
The quality the rewritten text from GPT-3.5 was validated on 100 individual resumes.
Although some sentences were rephrased slightly, the rewritten text was found to be accurate overall.

\paragraph{Enrich with Occupation Labels:} Every experience in our dataset is enriched with its corresponding occupation out of all 3007 ESCO occupations available.
We use a proprietary classifier that is able to accurately classify each experience based on its title and description.
An extensive manual validation process on 10\% of the dataset confirmed the accuracy of these labels as only 2.2\% of labels were found to be suboptimal.
These ESCO labels are stored as part of the final dataset.
Note that the 3007 ESCO occupations do not capture all aspects of the roles, as they for example do not reflect different seniority levels within a role.
Rather, they provide a high-level categorisation of jobs based on their performed activities.

\subsection{Dataset Analysis}
\label{sec:analytics}

\begin{figure}[ht]
\centering
\includegraphics[width=\columnwidth]{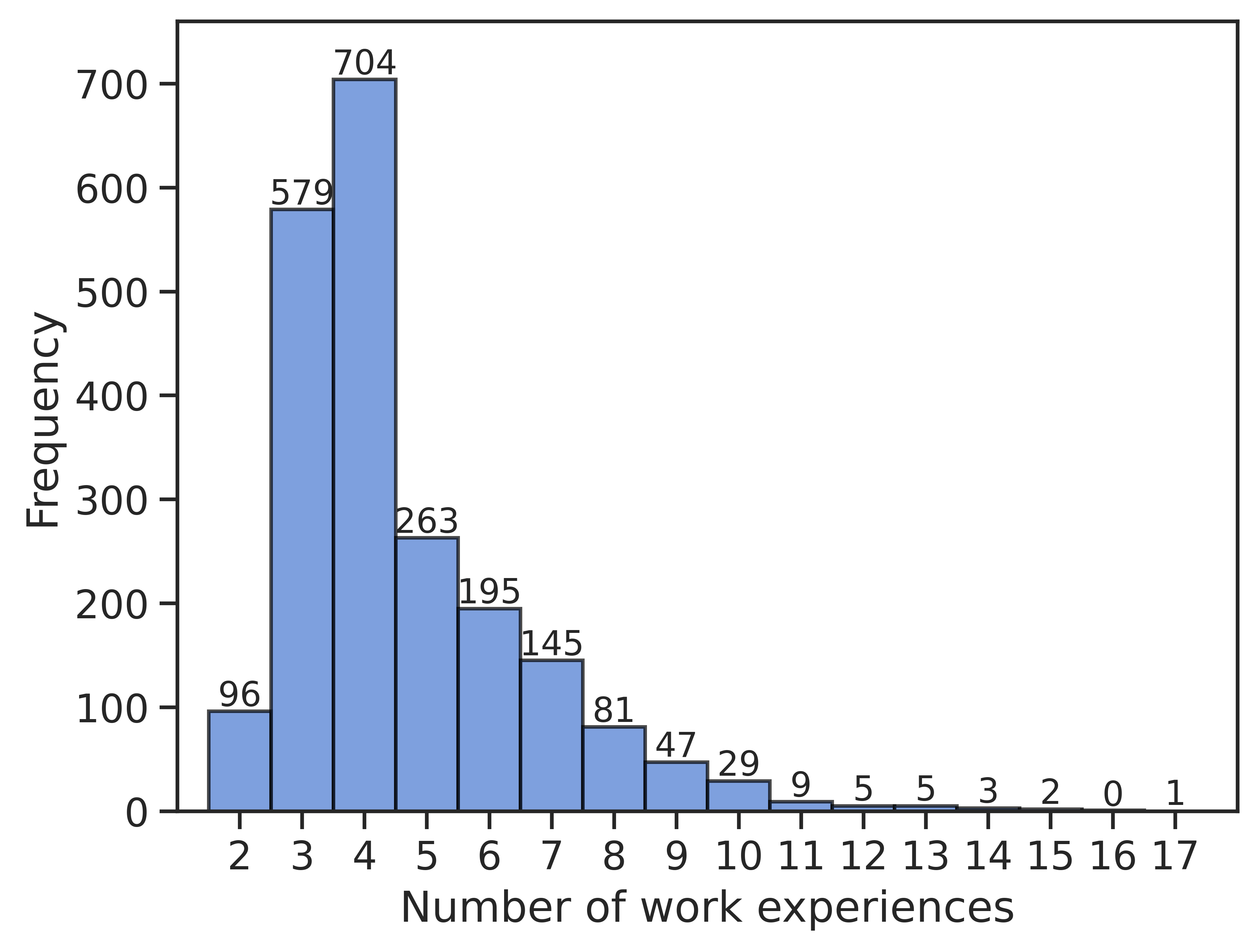}
\caption{Histogram of the number of work experiences per resume in our dataset.}
\label{fig:histogram}
\end{figure}

The industries are relatively balanced across the dataset, with 18 out of 24 industries having between 90 to 108 resumes.
A detailed breakdown is included in \cref{app:breakdown}.
\Cref{fig:histogram} shows the distribution of the number of experiences per career history.

The ESCO occupations in our dataset follow a long-tailed distribution, as can be seen in detail from the $\log$-$\log$ plot in \cref{app:breakdown}.
The most frequent 300 ESCO occupations represent a little over 80\% of all experiences in the dataset, while over 60\% of ESCO occupations never appear in the dataset.

\section{Career Path Prediction Models}
\label{sec:models}
\subsection{Task Description}

\begin{figure*}[ht]
\centering
\includegraphics[width=\textwidth]{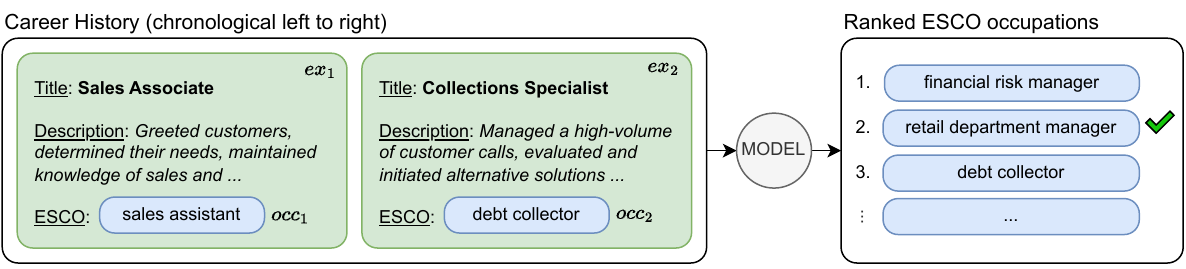}
\caption{\setlength\fboxsep{1pt}High-level illustration of the task: given a career history, rank all 3007 ESCO occupations in order of how suitable they are as next step. The career history is a chronological list of \colorbox{sentencecolor}{work experiences} and their \colorbox{descriptioncolor}{ESCO occupation} labels. The ESCO occupation of the left out next role in the career history (indicated by \faCheck) serves as round truth label and its rank is used for the rank-based evaluation metrics.}
\label{fig:overview}
\end{figure*}

We formalize career path prediction on our dataset as ranking the full set of ESCO occupations by how suitable they are as a next career step, based on the career history up until then, as illustrated in \cref{fig:overview}.
Each career history $(\experience{1}, \ldots, \experience{N})$ corresponds to $N - 1$ different prediction problems: for each experience $\experience{i}$ except the first one, its corresponding ESCO occupation label $\esco{i}$ serves as the true label to predict based on the preceding $i-1$ experiences.
More formally, we expect a scoring function $S((\experience{1}, \ldots, \experience{i-1}), \esco{})$ that takes a sequence of experiences and any ESCO occupation $\esco{}$ and outputs a score, after which all ESCO occupations are scored against the experience history $(\experience{1}, \ldots, \experience{i-1})$, and ranked from high to low scores.
The highest scored ESCO label should be the true label $\esco{i}$.
However, applications that rank recommended jobs to candidates can typically show more than one recommended job.
As such, we use rank-based metrics with a focus on top 5 and top 10 ranked occupations, specifically Mean Reciprocal Rank (MRR), recall@5 (R@5) and recall@10 (R@10).

To solve the ranking problem, in \cref{sec:ontologybased} we detail approaches that use the information contained within the ESCO ontology.
Next, \cref{sec:representationlearning} presents a combination of representation learning and regression to tackle the problem.
Finally, \cref{sec:hybrid} describes a hybrid method combining both.

\subsection{Skill-based Prediction}
\label{sec:ontologybased}

We hypothesize that job positions taken strongly rely on the skills of the person, and thus intuitively expect that the career path prediction could benefit from information on underlying skills. 
Such information is inherently present in ESCO, which captures both skills and job titles.
As the inferred ESCO labels for all experiences are available, we can make use of the full ESCO ontology, its attributes and structure to predict next jobs.
In the ESCO ontology, each occupation $\esco{}$ is linked to a set of standardized skills, which is partitioned in skills that are either ``essential'' or ``optional'' for $\esco{}$.
We denote such unified skill set combining both essential and optional skills as $\mathcal{S}(\esco{})$.
Given a career history with ESCO occupation labels $\esco{1}, \ldots, \esco{N}$, we represent the skills of the full career as the union of all related skills $\bigcup_{i=1}^{N} \mathcal{S}(\esco{i})$.
Finally, as a score to rank potential ESCO occupations $\esco{}$, we define the \emph{skill match} $S_{\text{SKILLS}}$ of an experience history against a specific ESCO occupation as the fraction of skills linked to that ESCO occupation that are also present in the union of skills associated with the work experience' ESCO labels, \ie

\[
S_{\text{SKILLS}}((\experience{1}, \ldots, \experience{N}), \esco{}) = \frac{\left|\bigcup_{i=1}^{N} \mathcal{S}(\esco{i}) \cap \mathcal{S}(\esco{})\right|}{\left|\mathcal{S}(\esco{})\right|}
\]

\subsection{Description-based Prediction}
\label{sec:representationlearning}

Our second model relies on the textual descriptions present in the career histories.
Given a sufficiently strong text representation model, we argue that it should be possible to 
predict next roles based on what has been described in previous experiences.
Two steps are necessary for this model.
First, a strong domain-specific representation model needs to be developed to accurately represent career histories and ESCO occupations in the same space.
Second, a mapping needs to be learned from the representation of a career history to the representation of relevant \textit{next} ESCO occupations, through which the career path prediction task can be performed.

\paragraph{Career History Representation Learning}
To learn a powerful domain-specific representation model for career histories, we make use of the parallel information that is contained in our dataset.
For each work experience in the dataset, we have two textual descriptions, being
\begin{enumerate*}[(1)]
\item the self-reported title and experience description from the resume, and
\item the ESCO occupation title as well as its ``description'' field in the ESCO ontology.
\end{enumerate*}
Inspired by this parallel textual data, we adopt a contrastive learning strategy to finetune a sentence-transformer model (\textit{all-mpnet-base-v2})\footnote{\url{https://huggingface.co/sentence-transformers/all-mpnet-base-v2}} that was pretrained on over 1B English sentence pairs~\cite{reimers2019sentence}.
We make use of \textit{multiple negatives ranking loss} with in-batch negatives, as proposed by \cite{Henderson2017EfficientNL}.
This training procedure only requires positive pairs (doc1, doc2) of corresponding textual documents.
We format both an experience's self-reported job title and description and those for an ESCO occupation in the same way, to embed them each with the chosen sentence-transformer (where we add the \textit{``esco''} prefix only for ESCO roles):

\begin{tcolorbox}[fontupper=\footnotesize, colback=lightgray!20, boxrule=0.5pt, arc=4pt, boxsep=0pt, left=4pt, right=4pt, top=2pt, bottom=2pt, colframe=black, sharp corners]
(esco) role: <title>\\
description: <description>
\end{tcolorbox}

\begin{figure*}[ht]
\centering
\includegraphics[width=\textwidth]{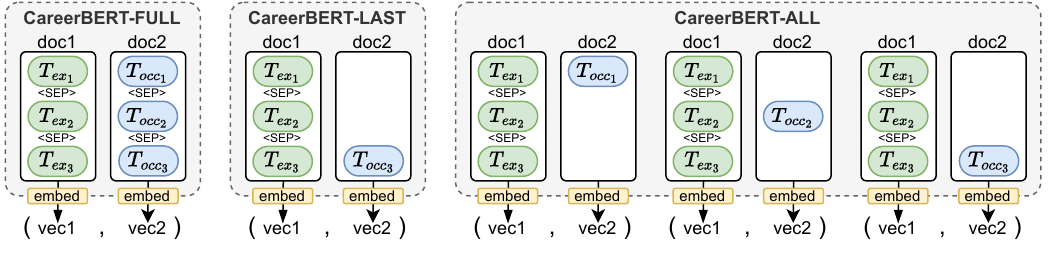}
\caption{\setlength\fboxsep{1pt}Illustration of the different strategies of creating positive pairs (doc1, doc2) for the contrastive training of \textsc{CareerBERT}.
The illustration considers a career history of three experiences, each with their \colorbox{sentencecolor}{self-reported} (left) and corresponding \colorbox{descriptioncolor}{ESCO occupation} (right) information.
Note that this applies to career history spans of any length.
The \textsc{CareerBERT-FULL} model uses pairs of completely corresponding sequences of self-reported and ESCO experiences.
\textsc{CareerBERT-LAST} on the other hand only retains the last ESCO experience.
Finally, \textsc{CareerBERT-ALL} is similar to \textsc{CareerBERT-LAST} but creates a text pair for each ESCO experience in the history.} \label{fig:trainingsetup}
\end{figure*}

Since we want to represent full career histories and not just individual work experiences,
multiple work experiences are combined in one document, by concatenating the single experience representations (ordering them chronologically from oldest to most recent), separated by the tokenizer's reserved \textsc{SEP} token, which we denote as $concat(T_{ex_1}, \cdots, T_{ex_N})$.
Now for each career trajectory, we want to create pairs (\textit{doc1}, \textit{doc2}) of textual representations of on the one hand the experiences as described in the resumes, and on the other hand the ESCO-ontology counterparts, to use in the contrastive training.
For this, we explore three different approaches (visualized in \cref{fig:trainingsetup}):
\begin{itemize}[leftmargin=*]
    \item \textbf{\textsc{CareerBERT-FULL}} -- given a career history, cast the sequence of self-reported experiences into \textit{doc1} and cast the corresponding sequence of ESCO occupations into \textit{doc2}.
    \item \textbf{\textsc{CareerBERT-LAST}} -- given a career history, cast the sequence of self-reported experiences into \textit{doc1} and cast only the \underline{last} ESCO occupation into \textit{doc2}.
    \item \textbf{\textsc{CareerBERT-ALL}} -- given a career history, cast the sequence of self-reported experiences into \textit{doc1}. For each ESCO occupation in the sequence, cast it separately into a \textit{doc2} text, generating as many pairs as the length of the sequence.
\end{itemize}

The \textsc{CareerBERT-FULL} is the typical scenario of contrastive learning in which we use two different (textual) representations of the same underlying information.
However, we suspect that this strategy might be limited in its effectiveness, as properties like the length of the text, or the amount of SEP tokens could already give away the correct matching of pairs within a batch, without considering the underlying meaning of the text.
To counter this expectation, the \textsc{CareerBERT-LAST} strategy is included.
This strategy uses only the last ESCO label in \textit{doc2}, thus avoiding the above mentioned risks.
However, a risk with this strategy is that the representation of the self-reported career history will focus only on the last part (the last experience).
A final strategy (\textsc{CareerBERT-ALL}) is thus included to counter this expectation.
This strategy is similar to \textsc{CareerBERT-LAST}, but duplicated for each ESCO label in the sequence instead of only the last one.
We hypothesize that, by \textit{doc2} randomly being one of the assigned ESCO labels, the representation of the self-reported career needs to be expressive of \underline{all} its experiences.

Finally, note that each contiguous subspan of a career history is a plausible career trajectory, and for each history with $N$ experiences, there exist $\frac{N \cdot (N + 1)}{2}$ such spans.
We use this insight the vastly increase the number of career trajectories that can be used in this representation learning stage.

\paragraph{Linear Projection}

As a second stage of the text-based career path prediction, a mapping needs to be learned from the career history representation to the representation of the next ESCO occupation. Formally, given a text representation function $embed$, we need to learn a mapping $P$ from $embed(concat(T_{ex_1}, \cdots, T_{ex_{N-1}}))$ to $embed(T_{occ_N})$.
While more sophisticated options are available, we take the simple approach of learning a linear transformation between both vectors, and optimize this using the ordinary least squares regression.
This projection $P$ then allows us to write down the text-based scoring function as follows:

\begin{multline*}
S_{\text{TEXT}}((ex_1, ex_2, \ldots, ex_N), occ) \\
= 
\text{cosim}(\,P(embed(concat(T_{ex_1}, \cdots, T_{ex_N})))
\,,\, embed(T_{occ})\,)
\end{multline*}
with 
\[
\text{cosim}(A,B) \triangleq \frac{A \cdot B}{\|A\| \cdot \|B\|}
\]

\subsection{Hybrid Prediction}
\label{sec:hybrid}

Finally, we combine the above metrics $S_\text{SKILL}$ and $S_\text{TEXT}$ because we hypothesize that the signal of skill-based prediction and description-based prediction are complementary.
Introducing just one hyperparamter $\alpha$, our hybrid approach is defined as the weighted sum:
\[
S_{\text{HYBRID}} = \alpha \cdot S_{\text{TEXT}} + (1 - \alpha) \cdot S_{\text{SKILL}}.
\]

\section{Experimental results and Discussion}
\label{sec:results}

We split our dataset randomly into a train, validation and test subset (80\%/10\%/10\%), stratified along the industries to maintain diverse profiles in each.
The statistics of each subset are shown in \cref{table:stats}.

\begin{table}[h]
\centering
\begin{tabular}{lcc}
\hline
& Career Histories & Experiences \\
\hline
Train & 1720 & 7912 \\
Validation & 217 & 957 \\
Test & 227 & 1050 \\
\hline
\end{tabular}
\caption{Statistics of the train, validation and test subsets of the dataset.}
\label{table:stats}
\end{table}

The different \textsc{CareerBERT} models are trained on the train subset, for a maximum of 2 epochs.
During training, we measure the loss on the validation set every 10\% of an epoch, and keep the best performing checkpoint.
We refer to \cref{app:trainingdetails} for further details about the training procedure.
In the rest of this section, we first validate the quality of each \textsc{CareerBERT} strategy through the industry classification task in \cref{sec:repquality}.
Then the main task of career path prediction is evaluated in \cref{sec:cpp}.

\subsection{Representation Learning Quality}\label{sec:repquality}

An initial validation of the \textsc{CareerBERT} representation models is performed as to better understand and compare their effectiveness in representing career histories.
For this, we use the industry classification task as proposed in \cite{inoubli2022dgl4c}.
Each career history in our dataset is linked to one in 24 total industries.
The quality of the representation model, when kept frozen and combined with a simple classification layer, should correlate with performance on this prediction task.
We follow the same setup as \cite{inoubli2022dgl4c} which is to sample 80\% of all histories for training and the other 20\% for validation.
This is measured across 10 different random splits.
We use a one-vs-all support-vector machine (SVM) for the classification.
\Cref{table:reprqualres} shows the average accuracy across the 10 random runs, as well as their standard deviations.
The pretrained model without any finetuning is included for reference.
We observe that \textsc{CareerBERT-ALL} leads to the highest performance in this case.

\begin {table}[ht]
\begin{tabular}{lccc}\toprule
                                   & Accuracy (\%) \\\midrule
Pretrained                         & 61.82 \textpm 1.70 \\
CareerBERT-FULL                    & \underline{67.14 \textpm 1.72} \\
CareerBERT-LAST                    & 66.40 \textpm 1.37 \\
CareerBERT-ALL                     & \textbf{68.94 \textpm 1.70} \\

\bottomrule
\end{tabular}
\caption{Average industry classification accuracy and standard deviation across 10 runs, for each \textsc{CareerBERT} strategy.}
\label{table:reprqualres}
\end{table}

\subsection{Career Path Prediction}\label{sec:cpp}

\begin{figure}[ht]
\centering
\includegraphics[width=0.5\textwidth]{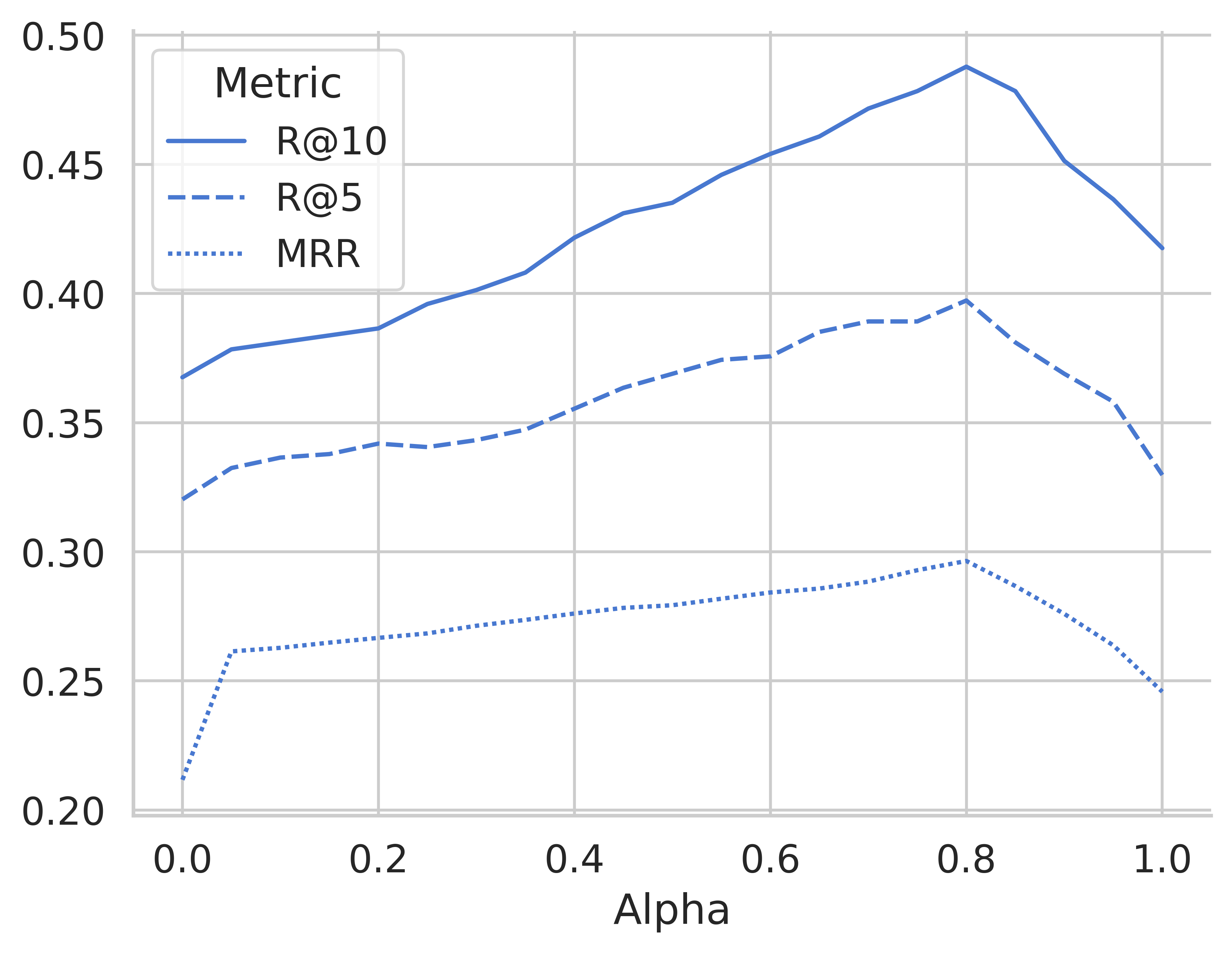}
\caption{Grid search for the optimal $\alpha$ value, measured on the validation set. Completely on the left represents a full reliance on skill-based prediction, while completely on the right represents full text-based prediction using the \textsc{CareerBERT-ALL}\textsubscript{proj} method. The optimal value is observed at $\alpha=0.8$.} \label{fig:gridsearch}
\end{figure}

We include a simple baseline system ``reversed history'' which simply predicts the ESCO occupations present in the input, ranked most to least recent.
Our formulation of skill-based career path prediction has no parameters that can be tuned, so we directly report performance on the test set.
For the text-based prediction, no hyperparameter needs to be tuned.
Therefore, for each \textsc{CareerBERT} strategy, we directly train the linear projection on the combined train and validation set to report performance on the test set.
We include the pretrained encoder model without any finetuning for comparison.
Also, for each text representation model, we measure rank-based results with and without the linear projection, to estimate the impact of this stage.
Finally, for the hybrid prediction method, the $\alpha$ parameter needs to be tuned.
We perform a grid search for values between 0 and 1 with increments of 0.1 and measure performance for each value on the validation set, as shown in \cref{fig:gridsearch}.
As text-based method for this grid search, we decide to use the \textsc{CareerBERT-ALL} method as it seems to perform favorably.
The projection in this case is optimized on just the train set, as to not overfit on the validation set for this grid search.
Based on this grid search, the value for $\alpha$ was set to 0.8 for best results.
All results on the test set are compiled in table~\cref{table:fullhistorytest}.

\begin {table}[ht]
\begin{tabular}{lccc}\toprule
                                    & MRR    & R@5   & R@10  \\\midrule
\multicolumn{4}{c}{\centering \textbf{Baseline}} \\\midrule
Reverse history                     & \textbf{0.211} & \textbf{26.37} & \textbf{26.49} \\\midrule
\multicolumn{4}{c}{\centering \textbf{Skill-based Prediction}} \\\midrule
Skill-based prediction              & \textbf{0.211} & \textbf{29.04} & \textbf{35.24} \\\midrule
\multicolumn{4}{c}{\centering \textbf{Text-based Prediction}} \\\midrule
Pretrained                          & 0.168 & 26.73 & 34.99 \\
Pretrained\textsubscript{proj}      & 0.202 & 26.85 & 34.63 \\
CareerBERT-FULL                     & 0.214 & 29.89 & 35.97 \\
CareerBERT-FULL\textsubscript{proj} & 0.232 & 31.59 & 36.94 \\
CareerBERT-LAST                     & 0.220 & 30.98 & 39.25 \\
CareerBERT-LAST\textsubscript{proj} & \underline{0.233} & \underline{31.96} & 38.52 \\
CareerBERT-ALL                      & 0.200 & 29.16 & \textbf{39.61} \\
CareerBERT-ALL\textsubscript{proj}  & \textbf{0.247} & \textbf{32.44} & \underline{39.49} \\\midrule
\multicolumn{4}{c}{\centering \textbf{Hybrid Prediction}} \\\midrule
$\alpha=0.8$                        & \textbf{0.274} & \textbf{37.06} & \textbf{43.01} \\
\bottomrule
\end{tabular}
\caption{Final performance of all methods on the test set. The strongest results in each prediction method are shown in bold, and second-best results (when applicable) are underlined.}
\label{table:fullhistorytest}
\end{table}

We observe that the baseline using reverse history reaches 26.37\% recall@5 and only 26.49\% recall@10, which reflects the limited information available in this simple baseline.
The skill-based prediction method surpasses the baseline with close to 9 \%-points recall@10.
Among the text-based prediction methods, we observe that \textsc{CareerBERT-ALL} performs strongest.
This validates our assumption that stronger representation models (as represented on the industry classification task) indeed lead to stronger results for career path prediction as well.
Adding the linear projection increases performance in general, although recall@10 seems to go down a bit in some cases.
Finally, we show that skill-based and text-based prediction are complementary, as the hybrid approach reaches the overall best results on all metrics.

\section{Conclusion and Future Work}

We develop and release a new dataset of over 2,164 anonymous work histories annotated with ESCO occupations.
The dataset is unique in its focus on the free-form textual descriptions that come with work experiences in resumes.
Through this dataset, we formulated \textsc{CareerBERT}, a novel representation learning technique tailored for work history texts.
We study different approaches to train \textsc{CareerBERT} and find non-trivial quality differences.
The strongest performance for both industry classification and career path prediction is obtained using the \textsc{CareerBERT-ALL} strategy, which is in line with our expectations when designing the different strategies.
Our research yielded two distinct models: a skill-based and a text-based model for career path prediction.
Next to the textual information, underlying skills and the match between current skills and skills for future jobs plays an important role.
Combining both text-based and skill-based predictions turns out to work best due to their information being complementary.\\
We left out the period and duration of work experiences from our experiments, but this would be interesting to include in future work.
Furthermore, future work might investigate how more of the structured information in the ESCO ontology could be leveraged to increase the performance of career path prediction even more.

\begin{acknowledgments}
We thank the anonymous reviewers for their valuable feedback. This project was funded by the Flemish Government, through Flanders Innovation \& Entrepreneurship (VLAIO, project HBC.2020.2893).
\end{acknowledgments}

\newpage

\bibliography{paper.bib}

\begin{thebibliography}{16}
\expandafter\ifx\csname natexlab\endcsname\relax\def\natexlab#1{#1}\fi
\providecommand{\url}[1]{\texttt{#1}}
\providecommand{\href}[2]{#2}
\providecommand{\path}[1]{#1}
\providecommand{\DOIprefix}{doi:}
\providecommand{\ArXivprefix}{arXiv:}
\providecommand{\URLprefix}{URL: }
\providecommand{\Pubmedprefix}{pmid:}
\providecommand{\doi}[1]{\href{http://dx.doi.org/#1}{\path{#1}}}
\providecommand{\Pubmed}[1]{\href{pmid:#1}{\path{#1}}}
\providecommand{\bibinfo}[2]{#2}
\ifx\xfnm\relax \def\xfnm[#1]{\unskip,\space#1}\fi
\bibitem[{Iqbal et~al.(2012)Iqbal, Latif, and Naseer}]{iqbal2012impact}
\bibinfo{author}{M.~T. Iqbal}, \bibinfo{author}{W.~Latif},
  \bibinfo{author}{W.~Naseer},
\newblock \bibinfo{title}{The impact of person job fit on job satisfaction and
  its subsequent impact on employees performance},
\newblock \bibinfo{journal}{Mediterranean Journal of Social Sciences}
  \bibinfo{volume}{3} (\bibinfo{year}{2012}) \bibinfo{pages}{523--530}.
\bibitem[{Chhabra(2015)}]{chhabra2015person}
\bibinfo{author}{B.~Chhabra},
\newblock \bibinfo{title}{Person--job fit: Mediating role of job satisfaction
  \& organizational commitment},
\newblock \bibinfo{journal}{The Indian Journal of Industrial Relations}
  (\bibinfo{year}{2015}) \bibinfo{pages}{638--651}.
\bibitem[{De~Neve et~al.(2018)De~Neve, Krekel, and Ward}]{de2018work}
\bibinfo{author}{J.-E. De~Neve}, \bibinfo{author}{C.~Krekel},
  \bibinfo{author}{G.~Ward},
\newblock \bibinfo{title}{Work and well-being: A global perspective},
\newblock \bibinfo{journal}{Global happiness policy report}
  (\bibinfo{year}{2018}) \bibinfo{pages}{74--128}.
\bibitem[{Zhang et~al.(2021)Zhang, Zhou, Zhu, Xu, Zha, Chen, and
  Xiong}]{zhang2021attentive}
\bibinfo{author}{L.~Zhang}, \bibinfo{author}{D.~Zhou},
  \bibinfo{author}{H.~Zhu}, \bibinfo{author}{T.~Xu}, \bibinfo{author}{R.~Zha},
  \bibinfo{author}{E.~Chen}, \bibinfo{author}{H.~Xiong},
\newblock \bibinfo{title}{Attentive heterogeneous graph embedding for job
  mobility prediction},
\newblock in: \bibinfo{booktitle}{Proceedings of the 27th ACM SIGKDD conference
  on knowledge discovery \& data mining}, \bibinfo{year}{2021}, pp.
  \bibinfo{pages}{2192--2201}.
\bibitem[{Yamashita et~al.(2022)Yamashita, Li, Tran, Zhang, and
  Lee}]{yamashita2022looking}
\bibinfo{author}{M.~Yamashita}, \bibinfo{author}{Y.~Li},
  \bibinfo{author}{T.~Tran}, \bibinfo{author}{Y.~Zhang},
  \bibinfo{author}{D.~Lee},
\newblock \bibinfo{title}{Looking further into the future: Career pathway
  prediction},
\newblock \bibinfo{journal}{WSDM Computational Jobs Marketplace 2022}
  (\bibinfo{year}{2022}).
\bibitem[{ESCO(2017)}]{ESCO}
\bibinfo{author}{ESCO},
\newblock \bibinfo{title}{European skills, competences, qualifications and
  occupations},
\newblock \bibinfo{journal}{EC Directorate E}  (\bibinfo{year}{2017}).
\bibitem[{Zhang et~al.(2019)Zhang, Liu, Zhu, Liu, Wang, Wang, and
  Xiong}]{zhang2019job2vec}
\bibinfo{author}{D.~Zhang}, \bibinfo{author}{J.~Liu}, \bibinfo{author}{H.~Zhu},
  \bibinfo{author}{Y.~Liu}, \bibinfo{author}{L.~Wang},
  \bibinfo{author}{P.~Wang}, \bibinfo{author}{H.~Xiong},
\newblock \bibinfo{title}{Job2vec: Job title benchmarking with collective
  multi-view representation learning},
\newblock in: \bibinfo{booktitle}{Proceedings of the 28th ACM International
  Conference on Information and Knowledge Management}, \bibinfo{year}{2019},
  pp. \bibinfo{pages}{2763--2771}.
\bibitem[{Xu et~al.(2014)Xu, Li, Gupta, Bugdayci, and Bhasin}]{xu2014modeling}
\bibinfo{author}{Y.~Xu}, \bibinfo{author}{Z.~Li}, \bibinfo{author}{A.~Gupta},
  \bibinfo{author}{A.~Bugdayci}, \bibinfo{author}{A.~Bhasin},
\newblock \bibinfo{title}{Modeling professional similarity by mining
  professional career trajectories},
\newblock in: \bibinfo{booktitle}{Proceedings of the 20th ACM SIGKDD
  international conference on Knowledge discovery and data mining},
  \bibinfo{year}{2014}, pp. \bibinfo{pages}{1945--1954}.
\bibitem[{Maheshwary and Misra(2018)}]{maheshwary2018matching}
\bibinfo{author}{S.~Maheshwary}, \bibinfo{author}{H.~Misra},
\newblock \bibinfo{title}{Matching resumes to jobs via deep siamese network},
\newblock in: \bibinfo{booktitle}{Companion Proceedings of the The Web
  Conference 2018}, \bibinfo{year}{2018}, pp. \bibinfo{pages}{87--88}.
\bibitem[{Lavi et~al.(2021)Lavi, Medentsiy, and Graus}]{lavi2021consultantbert}
\bibinfo{author}{D.~Lavi}, \bibinfo{author}{V.~Medentsiy},
  \bibinfo{author}{D.~Graus},
\newblock \bibinfo{title}{consultantbert: Fine-tuned siamese sentence-bert for
  matching jobs and job seekers},
\newblock \bibinfo{journal}{arXiv preprint arXiv:2109.06501}
  (\bibinfo{year}{2021}).
\bibitem[{Liu et~al.(2016)Liu, Zhang, Nie, Yan, and Rosenblum}]{liu2016fortune}
\bibinfo{author}{Y.~Liu}, \bibinfo{author}{L.~Zhang}, \bibinfo{author}{L.~Nie},
  \bibinfo{author}{Y.~Yan}, \bibinfo{author}{D.~Rosenblum},
\newblock \bibinfo{title}{Fortune teller: predicting your career path},
\newblock in: \bibinfo{booktitle}{Proceedings of the AAAI conference on
  artificial intelligence}, volume~\bibinfo{volume}{30}, \bibinfo{year}{2016}.
\bibitem[{Li et~al.(2017)Li, Jing, Tong, Yang, He, and Chen}]{li2017nemo}
\bibinfo{author}{L.~Li}, \bibinfo{author}{H.~Jing}, \bibinfo{author}{H.~Tong},
  \bibinfo{author}{J.~Yang}, \bibinfo{author}{Q.~He}, \bibinfo{author}{B.-C.
  Chen},
\newblock \bibinfo{title}{Nemo: Next career move prediction with contextual
  embedding},
\newblock in: \bibinfo{booktitle}{Proceedings of the 26th International
  Conference on World Wide Web Companion}, \bibinfo{year}{2017}, pp.
  \bibinfo{pages}{505--513}.
\bibitem[{Inoubli and Brun(2022)}]{inoubli2022dgl4c}
\bibinfo{author}{W.~Inoubli}, \bibinfo{author}{A.~Brun},
\newblock \bibinfo{title}{Dgl4c: a deep semi-supervised graph representation
  learning model for resume classification}  (\bibinfo{year}{2022}).
\bibitem[{Bhoomika et~al.(2023)Bhoomika, Likhitha, Chandana, Kavya, and
  Bhargavi}]{10169980}
\bibinfo{author}{S.~Bhoomika}, \bibinfo{author}{S.~Likhitha},
  \bibinfo{author}{H.~S. Chandana}, \bibinfo{author}{S.~A. Kavya},
  \bibinfo{author}{K.~Bhargavi},
\newblock \bibinfo{title}{2q-learning scheme for resume screening},
\newblock in: \bibinfo{booktitle}{2023 4th International Conference for
  Emerging Technology (INCET)}, \bibinfo{year}{2023}, pp.
  \bibinfo{pages}{1--5}. \DOIprefix\doi{10.1109/INCET57972.2023.10169980}.
\bibitem[{Reimers and Gurevych(2019)}]{reimers2019sentence}
\bibinfo{author}{N.~Reimers}, \bibinfo{author}{I.~Gurevych},
\newblock \bibinfo{title}{Sentence-{BERT}: Sentence embeddings using {S}iamese
  {BERT}-networks},
\newblock in: \bibinfo{booktitle}{Proceedings of the 2019 Conference on
  Empirical Methods in Natural Language Processing and the 9th International
  Joint Conference on Natural Language Processing (EMNLP-IJCNLP)},
  \bibinfo{publisher}{Association for Computational Linguistics},
  \bibinfo{address}{Hong Kong, China}, \bibinfo{year}{2019}, pp.
  \bibinfo{pages}{3982--3992}. \DOIprefix\doi{10.18653/v1/D19-1410}.
\bibitem[{Henderson et~al.(2017)Henderson, Al-Rfou, Strope, Sung, Luk{\'a}cs,
  Guo, Kumar, Miklos, and Kurzweil}]{Henderson2017EfficientNL}
\bibinfo{author}{M.~Henderson}, \bibinfo{author}{R.~Al-Rfou},
  \bibinfo{author}{B.~Strope}, \bibinfo{author}{Y.-H. Sung},
  \bibinfo{author}{L.~Luk{\'a}cs}, \bibinfo{author}{R.~Guo},
  \bibinfo{author}{S.~Kumar}, \bibinfo{author}{B.~Miklos},
  \bibinfo{author}{R.~Kurzweil},
\newblock \bibinfo{title}{Efficient natural language response suggestion for
  smart reply},
\newblock \bibinfo{journal}{ArXiv} \bibinfo{volume}{abs/1705.00652}
  (\bibinfo{year}{2017}).

\end{thebibliography}

\appendix

\section{Dataset Details}\label{app:breakdown}

\Cref{table:industries} shows all industries present in the dataset, with their number of career histories and average number of roles in those histories attached.

\begin{table}[h]
  \centering
  \caption{Industry Distribution}
  \begin{tabular}{lcc}
    \toprule
    Industry & Count & Average Roles \\
    \midrule
    FINANCE & 108 & 4.46 \\
    SALES & 107 & 4.32 \\
    ACCOUNTANT & 106 & 4.45 \\
    BUSINESS-DEVELOPMENT & 106 & 4.79 \\
    ADVOCATE & 104 & 4.88 \\
    CHEF & 103 & 4.95 \\
    CONSULTANT & 103 & 4.59 \\
    FITNESS & 102 & 4.57 \\
    IT & 102 & 4.02 \\
    PUBLIC-RELATIONS & 99 & 4.73 \\
    BANKING & 98 & 4.38 \\
    HR & 98 & 4.29 \\
    HEALTHCARE & 98 & 4.84 \\
    ENGINEERING & 97 & 4.29 \\
    ARTS & 93 & 4.24 \\
    AVIATION & 92 & 3.84 \\
    TEACHER & 91 & 4.34 \\
    DESIGNER & 90 & 4.94 \\
    CONSTRUCTION & 88 & 4.25 \\
    APPAREL & 87 & 5.76 \\
    DIGITAL-MEDIA & 82 & 5.04 \\
    AGRICULTURE & 62 & 4.74 \\
    AUTOMOBILE & 29 & 5.45 \\
    BPO & 19 & 4.95 \\
    \bottomrule
  \end{tabular}
\label{table:industries}
\end{table}

\newpage

A logarithmic plot of all ESCO occupation frequencies in the dataset is shown in \cref{fig:longtail} below.

\begin{figure}[ht]
\centering
\includegraphics[width=\columnwidth]{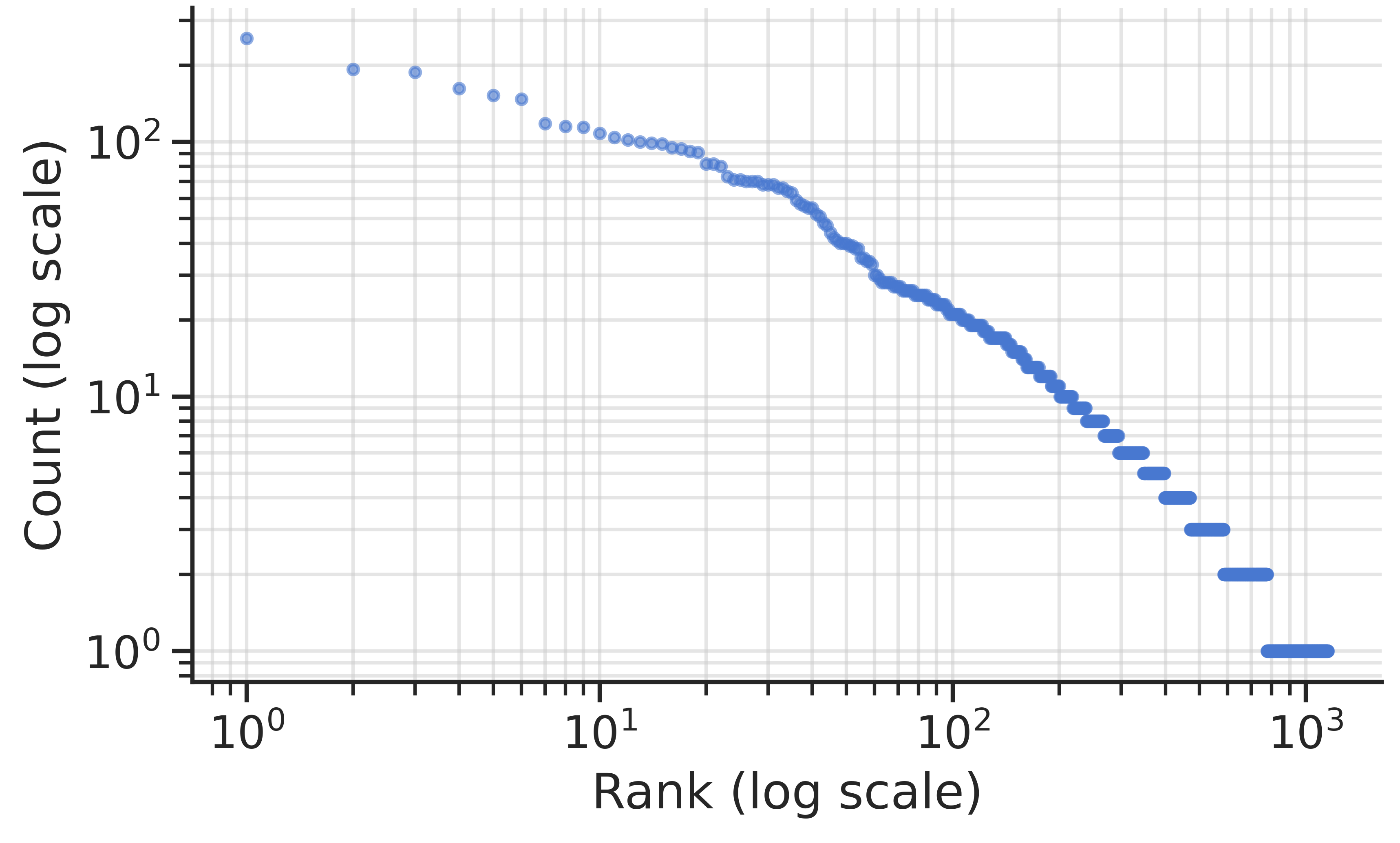}
\caption{Log-log plot of ESCO occupation frequencies in our career history dataset.} \label{fig:longtail}
\end{figure}

\section{CareerBERT Training Details}\label{app:trainingdetails}

The contrastive training is implemented using the popular SBERT implementation~\cite{reimers2019sentence}.
We keep the default value of 20 for the ``scale'' hyperparameter \textit{alpha}.
The positive pairs are randomly shuffled into batches of 16.
We use the AdamW optimizer with a learning rate of 2e-5 and a ``WarmupLinear'' learning rate schedule with a warmup period of 5\% of the training data.
Automatic mixed precision was used to speed up training.
All experiments where performed using an Nvidia T4 GPU.

\section{GPT-3.5 Prompt For Experience Reformatting}
\label{app:gpt-prompt-for-reformatting}

Below, the exact prompt used to rewrite the working histories is shown.
The prompt makes use of the conversational interface of the GPT-3.5 model, and consists of only one user message.
The position in which the original text is inserted is indicated in the prompt with \underline{text}.

\begin{tcolorbox}[fontupper=\footnotesize, colback=lightgray!20, boxrule=0.5pt, arc=4pt, boxsep=0pt, left=4pt, right=4pt, top=2pt, bottom=2pt, colframe=black, sharp corners]
\texttt{\textbf{User:}
\#\# Resume\\
\\
\underline{text}\\
\\
\#\# Task\\
\\
Rewrite the working history with the following format:\\
Role: <role>\\
Start: <start>\\
End: <end>\\
Description: <description>}
\end{tcolorbox}

\end{document}